\documentclass[sigconf]{acmart}
\usepackage{booktabs} % For formal tables
\usepackage{algorithm}
\usepackage{algorithmic}

\AtBeginDocument{%
  \providecommand\BibTeX{{%
    \normalfont B\kern-0.5em{\scshape i\kern-0.25em b}\kern-0.8em\TeX}}}

\copyrightyear{2019}
\acmYear{2019}
\acmConference[CIKM '19]{The 28th ACM International Conference on Information and Knowledge Management}{November 3--7, 2019}{Beijing, China}
\acmBooktitle{The 28th ACM International Conference on Information and Knowledge Management (CIKM '19), November 3--7, 2019, Beijing, China}
\acmPrice{15.00}
\acmDOI{10.1145/3357384.3358075}
\acmISBN{978-1-4503-6976-3/19/11}

\begin{document}
\fancyhead{}

\title{MarlRank: Multi-agent Reinforced Learning to Rank}

\author{Shihao Zou}
\authornote{This work was done when Shihao was with UCL and interned at Huawei.}
\email{szou2@ualberta.ca}
\affiliation{%
  \institution{University of Alberta}
}

\author{Zhonghua Li}
\email{lizhonghua3@huawei.com}
\affiliation{%
  \institution{Huawei}}

\author{Mohammad Akbari}
\email{m.akbari@ucl.ac.uk}
\affiliation{%
  \institution{University College London}}
  
\author{Jun Wang}
\email{jun.wang@cs.ucl.ac.uk}
\affiliation{%
  \institution{University College London}}
  
\author{Peng Zhang}
\email{pzhang@tju.edu.cn}
\affiliation{%
  \institution{Tianjin University}}


\begin{abstract}
  When estimating the relevancy between a query and a document, ranking models largely neglect the mutual information among documents. A common wisdom is that if two documents are similar in terms of the same query, they are more likely to have similar relevance score. To mitigate this problem, in this paper, we propose a multi-agent reinforced ranking model, named MarlRank. In particular, by considering each document as an agent, we formulate the ranking process as a multi-agent Markov Decision Process (MDP), where the mutual interactions among documents are incorporated in the ranking process. To compute the ranking list, each document predicts its relevance to a query considering not only its own query-document features but also its similar documents' features and actions. By defining reward as a function of NDCG, we can optimize our model directly on the ranking performance measure. Our experimental results on two LETOR benchmark datasets show that our model has significant performance gains over the state-of-art baselines. We also find that the NDCG shows an overall increasing trend along with the step of interactions, which demonstrates that the mutual information among documents helps improve the ranking performance.
\end{abstract}

\begin{CCSXML}
<ccs2012>
<concept>
<concept_id>10002951.10003317.10003338.10003343</concept_id>
<concept_desc>Information systems~Learning to rank</concept_desc>
<concept_significance>500</concept_significance>
</concept>
<concept>
<concept_id>10002951.10003317.10003338.10010403</concept_id>
<concept_desc>Information systems~Novelty in information retrieval</concept_desc>
<concept_significance>500</concept_significance>
</concept>
</ccs2012>
\end{CCSXML}

\ccsdesc[500]{Information systems~Learning to rank}
\ccsdesc[500]{Information systems~Novelty in information retrieval}

\keywords{multi-agent reinforcement learning, learning to rank}


\maketitle
\section{Introduction}
Learning to rank, which learns a model for ranking documents (items) based on their relevance scores to a given query, is a key task in information retrieval (IR)~\cite{xu2007adarank}.  Typically, a query-document pair is represented as a feature vector and many classical models, including SVM~\cite{herbrich1999support}, AdaBoost~\cite{xu2007adarank} and regression tree~\cite{lambdamart2010burges} are applied to learn to predict the relevancy. Recently, deep learning models has also been applied in learning to rank task. For example, CDSSM~\cite{Shen2014learning},  DeepRank~\cite{Pang2017deeprank} and their variants aim at learning better semantic representation of the query-document pair for more accurate prediction of the relevancy. Basically, most of previous approaches emphasize on the learning models with only query-document features considered. They do not investigate the effectiveness of global mutual information among documents to improve the ranking performance, which will be the focus of this paper. 

Reinforcement learning has been applied in a few IR tasks, such as session search~\cite{Luo2014win,Zhang2014pomdp}, where a user's behavior in one session affects documents ranking in the next session. This can be formulated as an MDP and optimized using reinforcement learning. 
Recent works on reinforcement learning to rank model the ranking process as an MDP~\cite{wei2017reinforcement,feng2018from}, where, at each time step, an agent selects one document given its current observation (ranking position and un-ranked documents), and the reward is the improvement of NDCG. But essentially, this contextual information is limited as documents are passively selected and not allowed to interact with each other, and the interactions among documents can be considered as the mutual information among documents which helps to improve ranking performance.

Depart from prior approaches, we consider both query-document and document-document relevancy for ranking. Our hypothesis comes from a common wisdom that if two documents are very similar in terms of the same query, they are likely to have similar relevance scores. We give a motivational example to illustrate our intuition in Sec~\ref{explain_intuition}. Inspired by~\cite{lowe2017multi}, we construct the ranking process as an interactive multi-agent environment where each document is treated as an agent. 
% By representing each document as a query-document feature vector \cite{liu2007letor}, the similarity between documents is conditioned on the same query. Then the interactions among documents in the multi-agent ranking process include not only independent query-document information but also document-document mutual information. Besides, our model can be optimized directly on the ranking performance measure by multi-agent reinforcement learning. 
Specifically, our multi-agent environment comprises a query and a set of documents (agents). The ranking process is an interactive MDP among these agents. At each time step, each agent simultaneously presents its relevance score based on the observation of its own query-related information and the local information of its neighbor agents, and then a global ranking performance reward is given. After several time steps of interaction, all the agents achieve an equilibrium state. Their common objective is to find an optimal policy that jointly maximizes the expected accumulative reward. 

The contributions of this paper are as follows: (1) we propose to leverage both query-document and document-document relevancy for ranking in IR; (2) we formulate the learning to rank problem as a multi-agent reinforced learning framework. To the best of our knowledge, the proposed framework is the first work that models document ranking as an interactive multi-agent reinforced learning process and optimizes the model directly on the ranking performance measure of the retrieval task. The experimental results show that our model outperforms most strong baselines and it is interesting to find NDCG increases along with time steps, which attests the interactions among documents help improve the ranking performance.

\section{Background}
\subsection{Related Work}
Typically, ranking models can be classified into three categories: point-wise, pair-wise \cite{herbrich1999support,burges2005learning,burges2007learning} and list-wise \cite{cao2007learning,xu2007adarank,xu2008directly}. Point-wise ranking models $p(d|q;{\theta})$ assume that documents are independent of each other for predicting the relevance, where $\theta$ means the parameter of the ranking model. On the contrary, pair-wise models regard a pair of documents as an instance and the pair of documents' relevance orders as categories to perform classification-based ranking, referred as $p(d_i|q, d_j;\theta)$ where $d_i$ and $d_j$ are a pair of documents. Empirical results show that pair-wise ranking models outperform point-wise ones as, intuitively, pair-wise ranking process takes mutual information between a pair of documents in the ranking process. As for list-wise models, they formulate a ranking list (either a document pair \cite{xu2008directly} or a documents list \cite{cao2007learning}) as a permutation distribution so that loss functions become continuous and differentiable. Then list-wise models can be directly optimized on ranking performance measures. However, in essence, list-wise models mainly focus on the reformulation of loss functions and, like point-wise and pair-wise models, treat each document or each pair of documents dependently to do ranking. 

Traditional learning to rank models mainly focus on the minimization of classification errors, including RankSVM \cite{herbrich1999support}, RankBoost \cite{freund2003efficient} and RankNet \cite{burges2005learning}.
These models formulate ranking as a classification problem, either point-wise or pair-wise. Instead of optimizing the classification loss 
many previous works design a variety of differentiable loss functions which are directly based on the IR performance measure. For example, LambdaRank \cite{burges2007learning} speeds up RankNet by directly constructing the first derivative (lambda) of the implicit non-continuous cost (such as NDCG, MAP, etc.) with respect to the model parameters. Following this line, LambdaMART \cite{lambdamart2010burges} uses multiple additive regression tree, which shows better ranking performance. Another line of work is based on list-wise loss. ListNet \cite{cao2007learning} maps a list of document scores to a differentiable loss defined on the permutation probability distribution.
AdaRank \cite{xu2007adarank} revises the non-continuous list ranking accuracy loss to a continuous exponential upper bound. Besides, SVMNDCG \cite{chakrabarti2008structured} and PermuRank\cite{xu2008directly} directly optimize the upper bound defined on the pairs between the perfect and imperfect permutations. Our work departs from them in that our model uses ranking performance measure as the reward and we can optimize the model directly based on this ranking performance measure via policy gradient.

Reinforcement learning has shown great success in many tasks, including game \cite{silver2017mastering} and recommender system \cite{Zheng2018DRN}. In IR, there are mainly two lines of works which are formulated as an MDP. One line is session search. In \cite{Luo2014win,Zhang2014pomdp}, a collaborative search process between the user and the search engine is defined as an MDP.
The other line mainly focuses on reinforcement learning to rank. The ranking process is formulated as an MDP. \cite{wei2017reinforcement} At each time step, the agent selects one document in the ranking list until all the documents are selected and ranked. A further exploratory algorithm based on Monte Carlo tree search is proposed by \cite{feng2018from}. In addition, IRGAN \cite{wang2017irgan} uses policy gradient to optimize the generator. Our model is different from that of \cite{wei2017reinforcement,feng2018from} whose agent is only for the selection of documents. By regarding each document as an agent in a collaborative environment, we include mutual information among documents in our model to do ranking.

\subsection{A Motivational Example}\label{explain_intuition}
A motivational example is presented to illustrate how our intuition helps improve the ranking performance. In Fig. \ref{fig:toy example}, given a query, we assume there are six documents with three non-relevant $\{d_1, d_2, d_3\}$ and three relevant ones $\{d_4, d_5, d_6\}$. The initial predictions at step $0$ are given in Tab.~\ref{tab:toy example}, and we assume the naive policy is the average of the target document and its top-$2$ similar documents' previous actions (relevance scores). For example, the action of document $d_4$ at step $1$ is the average of previous actions by document $4,\ 5$ and $6$, $a_1^4=(a_0^4+a_0^5+a_0^6)/3=0.63$. We can see from Tab.~\ref{tab:toy example} that document $d_4$ finds its neighbor documents $d_5$ and $d_6$, so it believes that it should have similar scores with these two neighbors and updates its prediction from 0.1 to 0.63. Finally, NDCG@3 increases after one step of interaction and remains the best ranking performance afterwards.

\makeatletter\def\@captype{figure}\makeatother 
\begin{minipage}{0.36\columnwidth} 
    \vspace{0.5cm}
    \centering
    \includegraphics[scale=0.7]{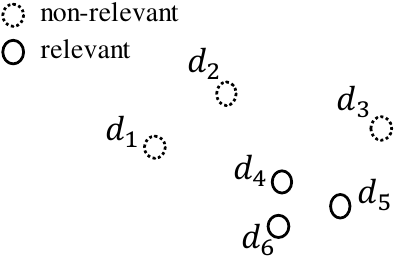}
    \setlength{\abovecaptionskip}{0pt}
    \setlength{\belowcaptionskip}{0pt}
    \caption{\small{Toy example.}}
    \label{fig:toy example}
\end{minipage}
\makeatletter\def\@captype{table}\makeatother 
\begin{minipage}{0.55\columnwidth} 
    \centering 
    \setlength{\abovecaptionskip}{0pt}
    \setlength{\belowcaptionskip}{0pt}
    \caption{Relevance score (action) given by each document.}
    \label{tab:toy example}
    \begin{tabular}{|c|c|c|c|c|} 
    \hline 
    step $t$     & 0 & 1    & 2    & 3  \\ 
    \hline
    $d_1\ (r=0)$ & 0 & 0.37 & 0.46 & 0.52 \\ 
    \hline
    $d_2\ (r=0)$ & 1.0 & 0.37 & 0.46 & 0.52 \\ 
    \hline
    $d_3\ (r=0)$ & 0 & 0.33 & 0.53 & 0.6 \\ 
    \hline
    $d_4\ (r=1)$ & 0.1 & 0.63 & 0.63 & 0.63 \\ 
    \hline
    $d_5\ (r=1)$ & 0.9 & 0.63 & 0.63 & 0.63 \\ 
    \hline
    $d_6\ (r=1)$ & 0.9 & 0.63 & 0.63 & 0.63 \\ 
    \hline
    \small{NDCG@3} & 0.3 & 1 & 1 & 1 \\ 
    \hline 
    \end{tabular} 
\end{minipage} 

\section{Framework}
\subsection{Reinforced Learning to Rank}
Given a query $q$, we have $N$ documents to rank, $\big(q, \{(d_i, y_i)\}_{i=1}^N\big)$, in which document $d_i$ is represented as a query-document feature vector as in \cite{liu2007letor} and $y_i$ is its relevance label. Now we formally define the Reinforced Learning to Rank problem as a multi-agent MDP of $\mathcal{M} = <\mathcal{S}, \mathcal{O}, \mathcal{A}, \mathcal{T}, \mathcal{R}, \pi, \gamma>$:
$\mathcal{S}=\big\{q, \{(d_i, y_i, a^i)\}_{i=1}^N\big\}$ is a set of environment states, where each document is regarded as an agent, and $a^i$ is the action of the $i$-th agent. $\mathcal{O}=\{\mathcal{O}^1,...,\mathcal{O}^N\}$ is a set of observations for each agent. As the environment is fully observable, the observation of each agent is equivalent to the environment state, $o_t^i=s_t$, where $o_t^i\in\mathcal{O}^i$ and $s_t\in\mathcal{S}$.

$\mathcal{A}=\{\mathcal{A}^1,...,\mathcal{A}^N\}$ is a set of actions for each agent. The action is defined as the discrete relevance level for the query $q$. For agent $i$, it uses the policy $\pi_{\theta}(a_t^i|o_t^i)$ to decide an action at time step $t$, where $\theta$ is the parameter of the policy $\pi$. To make it simple, we assume all the agents use the same policy to choose their actions given their own observations.

$\mathcal{T}:\mathcal{S}\times\mathcal{A}^1\times...\times\mathcal{A}^N\mapsto\mathcal{S}$ is the state transition function.

$\mathcal{R}:\mathcal{S}\times\mathcal{A}^1\times...\times\mathcal{A}^N\mapsto\mathbb{R}$ is the reward function, which is defined as
    \begin{equation}\label{eq:reward}
        r_t(a^1_t, ..., a^N_t) = \left\{
        \begin{aligned}
            & 0  & \quad t<T \\
            & NDCG_T - NDCG_{best} & \quad t=T
        \end{aligned}
        \right. ,
    \end{equation}
  where $T$ is the final time step and $NDCG_{best}$ means the best NDCG value, which is $1$ if no less than one document is relevant to the query and $0$ if no document is relevant. That is to say, only at the final step, a non-zero reward is given. The reason for our reward definition is given in Eq.~\eqref{eq:reward}.

The return at time step $t$ is defined as the accumulative reward in the future, $R_t=\sum_{k=t}^T\gamma^{k-t} r_k(a^1_k, ..., a^N_k)$. The objective is find an optimal policy to maximize the expected return:
\begin{equation}
\label{eq:objective}
    J(\theta) = \mathop{\arg\max}_{\theta} \mathbb{E}_{\pi}\Big[\sum_{k=t}^{T}\gamma^{k-t} r_k(a^1_k, ..., a^N_k)\Big],
\end{equation}
where $\gamma$ is the discounted factor and $T$ is the total time steps. Note that according to Eq.~\eqref{eq:reward}, the objective of Eq.~\eqref{eq:objective} becomes
\begin{align}
    J(\theta) &= \mathop{\arg\max}_{\theta} \mathbb{E}_{\pi}\Big[\gamma^{T-t}(NDCG_T - NDCG_{best})\Big]\nonumber\\
    &\propto \mathop{\arg\max}_{\theta} \mathbb{E}_{\pi}\big[NDCG_T\big],
\end{align}
which is consistent with the goal of learning to rank task.

\subsection{MarlRank}
\begin{figure}[htb]
    \centering
    \includegraphics[width=0.7\columnwidth]{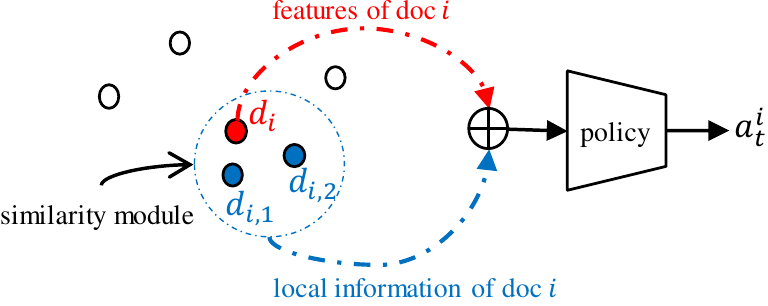}
    \setlength{\abovecaptionskip}{0pt}
    \setlength{\belowcaptionskip}{0pt}
    \caption{MarlRank: blue dashed circle is the similarity module, red and blue dashed arrow means doc $i$'s own features and its local information respectively. $d_{i,1}$ and $d_{i,2}$ are the top-$2$ similar documents of $d_i$.}
    \label{fig:model}
\end{figure}

The proposed model, MarlRank, includes similarity and policy two modules as is shown in Fig.~\ref{fig:model}. The similarity module is used to find the top-$k$ neighbor (similar) documents. We use $\big\{(d_{i,n}, s_{i,n}, a^{i,n}_{t-1})\big\}_{n=1}^k$ to denote the set of document $d_i$'s local information, where $s_{i,n}$ means the similarity between document $d_i$ and $d_n$ and $a^{i,n}_{t-1}$ means previous actions of neighbor document $d_n$. Then, we define the observation $o_t^i$ for $d_i$ at time step $t$ as
\begin{equation}
    o_t^i = concat\big[d_i,\ \{a_{t-1}^{i,n}\}_{n=1}^k,\ \{s_{i,n}\}_{n=1}^k,\  \frac{1}{k}\sum_{n=1}^k s_{i,n} d_{i,n} \big],
\end{equation}
whose first term is $d_i$'s own query-document feature vector and the other three terms are the local information of $d_i$, including previous actions, similarities and weighted sum of neighbor document's feature vector. Then given the observation $o_t^i$, the policy module decides an action $a_t^i$. To make it clear, we will use $\pi_{\theta}(a_t^i|o_t^i)$ to denote the two modules in the following part. 
% At each time step, all the documents simultaneously present their actions using the same policy. To make it simple, we will use $\pi_{\theta}$ to denote the entire model in the following part.

Besides, we also add a small individual reward $r_t^i(a_t^i, y_i)$ as a regularization term for each agent at time step $t$. The individual reward is positive when the agent's action equals to its label, and negative otherwise. Afterwards, we can update the model according to REINFORCE\cite{sutton1998introduction} as
\begin{equation}\label{eq:update}
    \nabla_{\theta} J(\theta) = \mathbb{E}_{\pi}\big[\nabla_{\theta}\log \pi_{\theta}(a_t^i|o_t^i)\ \big(R_t+r_t^i(a_t^i, y_i)\big)\big].
\end{equation}
Our algorithm is summarized in Algorithm~\ref{alg:MarlRank}.

% within one episode, a trajectory is sampled and a final non-zero global reward is given by the environment. Then we can compute the return $R_t=\sum_{k=t}^T\gamma^{k-t} r_k$ for each time step. 
%The overall logic of our proposed MarlRank is summarized in Algorithm~\ref{alg:MarlRank}.

\renewcommand{\algorithmicrequire}{ \textbf{Input:}}
\renewcommand{\algorithmicensure}{ \textbf{Output:}}
\begin{algorithm}[htb] 
    \caption{MarlRank} 
    \label{alg:MarlRank} 
    \begin{algorithmic}[1]
    \REQUIRE policy $\pi_{\theta}$; training dataset $D=\big\{\big(q, \{(d_i, y_i)\}_{i=1}^N\big)\big\}$\\
    \STATE Initialize $\pi_{\theta}$ with random weights $\theta$
    \STATE Pre-train $\pi_{\theta}$ using $D$
    \REPEAT
        \FOR{$\big(q, \{(d_i, y_i)\}_{i=1}^N\big)$ in $D$}
            \STATE Sample a trajectory $\{(o_t^i, a_t^i)\}$ using policy $\pi_{\theta}(a_t^i|o_t^i)$ \\
            \STATE Obtain a final reward by Eq.~\eqref{eq:reward} \\
            \STATE Compute return $R_t$ and individual reward $r_t^i(a_t^i, y_i)$ \\
        \ENDFOR
        \STATE Collect samples $\big\{\big(\ o_t^i,\ a_t^i,\ R_t+r_t^i(a_t^i, y_i)\ \big)\big\}$
        \STATE Update $\theta$ via REINFORCE according to Eq.~\eqref{eq:update}
    \UNTIL $\pi_{\theta}$ converges
    \end{algorithmic}
\end{algorithm}

\section{Experiments}
\subsection{Experiment setup}
We conduct experiments on two LETOR benchmark datasets \cite{liu2007letor}, MQ2007 and OHSUMED. Each dataset consists of queries, documents (query-document feature vectors) and relevance labels. Following LETOR, we conduct 5-fold cross-validation and calculate the average as the final test result. The relevance labels in both datasets consist of relevant ($2$), partially-relevant ($1$) and non-relevant ($0$). 

Our focus in this paper is on the effectiveness of global mutual information among documents to improve the ranking performance, rather than on the similarity models or sophisticated neural ranking models. So our model has only one-layer MLP for similarity module and two-layer MLP for policy module with each layer $100$ hidden units. The model outputs a probabilistic distribution over three relevance levels. We first pre-train our model to make it give better initial relevance prediction via supervised learning, and then train it via policy gradient. The individual reward used in OHSUMED is $0.001, 0.003, 0.004$ for the equal case of three relevance levels respectively and $-0.001$ for non-equal case, and $0.001, 0.003, 0.008$ and $-0.001$ in MQ2007. After one episode of sampling, the return is normalized to standard normal distribution. $\gamma$ is set to $0.95$ and learning rate is set to $4\times10^{-7}$ in both datasets. The initial previous action $a_{-1}^i$ is set to $0$ for all agents. In test and validation, the maximum time step is 10, though we find that the top-$10$ ranking order no longer changes after about $4$ steps for most queries. 

Several classical methods are used as baselines in Tab.~\ref{tab:mq2007} and \ref{tab:ohsumed}, whose results are obtained via RankLib\footnote{https://sourceforge.net/p/lemur/wiki/RankLib/}. To make a fair comparison, RankNet is the same three-layer MLP with each layer 100 hidden units as our model. Though LambdaMART has better implementations such as LightGBM, we should note that our model is a simple one and our focus is to prove that mutual information among documents is helpful to improve ranking.

\subsection{Experiment results}
Experimental results are shown in Tab.~\ref{tab:mq2007} and \ref{tab:ohsumed}. We can find that MarlRank outperforms most of baselines in terms of NDCG@1, 3, 5 and 10. In MQ2007, MarlRank has significant performance gains on NDCG@1, 5 and 10 over RankNet, AdaRank and ListNet. Our model also exceeds MDPRank on all four NDCG measures. This may lie in the fact that we include more mutual documents information in the ranking process. In OHSUMED, our model shows significant improvement on three baselines in terms of NDCG@1 and 10, while MDPRank is better on NDCG@3 and 10. The reason could be the much fewer amount of training data in OHSUMED than in MQ2007, which makes the model in OUSUMED perform a little worse than that in MQ2007. Fig.~\ref{fig:ndcg_along_with_time} shows how the NDCG measures change over time when evaluating MarlRank on both datasets. It is worth to notice from Fig.~\ref{fig:ndcg_along_with_time} that all four NDCG measures (computed after each time step on test set) increase along with the increase of time step for both datasets, which demonstrates that the interactions (mutual information) among documents help improve the ranking performance. While on OHSUMED all measures shows an overall increasing trend with slight fluctuations in the middle especially for NDCG@1,3, and 5, this is probably because of the imbalanced or inadequate  positive/negative documents for different queries.

\begin{table}
    \centering
    \setlength{\abovecaptionskip}{0pt}
    \setlength{\belowcaptionskip}{0pt}
    \caption{Test results on MQ2007. (*, $\sharp$, $\dagger$ and $\ddagger$ mean a significant improvement over AdaRank, RankNet, ListNet and LambdaMART using Wilcoxon signed-rank test $p<0.05$.)}
    \begin{tabular}{|c|cccc|} 
        \toprule
                  & \footnotesize{NDCG@1} & \footnotesize{NDCG@3} & \footnotesize{NDCG@5} & \footnotesize{NDCG@10}  \\ 
        \midrule
        \footnotesize{RankNet\cite{burges2005learning}}   & 0.3753 & 0.3802 & 0.3858 & 0.4173 \\ 
        \footnotesize{RankBoost\cite{freund2003efficient}} & 0.3968 & 0.3978 & 0.4046 & 0.4333 \\
        \footnotesize{AdaRank\cite{xu2007adarank}}   & 0.3751 & 0.3901 & 0.4013 & 0.4300 \\
        \footnotesize{ListNet\cite{cao2007learning}}   & 0.3870 & 0.3888 & 0.3936 & 0.4207 \\
        \footnotesize{LambdaMART\cite{lambdamart2010burges}}& 0.4154 & 0.4138 & \textbf{0.4197} & 0.4478 \\
        \footnotesize{MDPRank\cite{wei2017reinforcement}}   & 0.4033 & 0.4059 & 0.4113 & 0.4350 \\
        \midrule
        \footnotesize{MarlRank} & \textbf{0.4254}$^{*\sharp\dagger\ddagger}$ & \textbf{0.4139}$^{\sharp}$ & 0.4179$^{\sharp\dagger}$ & \textbf{0.4489}$^{*\sharp\dagger}$ \\
        \bottomrule
    \end{tabular} 
    \label{tab:mq2007}
    \vspace{-0.3cm}
\end{table}
\begin{table}
    \centering
    \setlength{\abovecaptionskip}{0pt}
    \setlength{\belowcaptionskip}{0pt}
    \caption{Test results on OHSUMED.}
    \begin{tabular}{|c|cccc|} 
        \toprule
                  & \footnotesize{NDCG@1} & \footnotesize{NDCG@3} & \footnotesize{NDCG@5} & \footnotesize{NDCG@10}  \\ 
        \midrule
        \footnotesize{RankNet\cite{burges2005learning}}   & 0.4667 & 0.4338 & 0.4356 & 0.4309 \\ 
        \footnotesize{RankBoost\cite{freund2003efficient}} & 0.5043 & 0.4739 & 0.4563 & 0.4385 \\
        \footnotesize{AdaRank\cite{xu2007adarank}}   & 0.5078 & \textbf{0.4913} & 0.4647 & 0.4478 \\
        \footnotesize{ListNet\cite{cao2007learning}}   & 0.4595 & 0.4325 & 0.4284 & 0.4181 \\
        \footnotesize{LambdaMART\cite{lambdamart2010burges}}& 0.4563 & 0.4523 & 0.4374 & 0.4299 \\
        \footnotesize{MDPRank\cite{wei2017reinforcement}}   & 0.5363 & 0.4885 & 0.4695 & \textbf{0.4591} \\
        \midrule
        \footnotesize{MarlRank} & \textbf{0.5521}$^{*\sharp\dagger\ddagger}$ & 0.4771$^{\sharp\dagger}$ & \textbf{0.4698}$^{\dagger}$ & 0.4551$^{\sharp\dagger\ddagger}$ \\
        \bottomrule
    \end{tabular} 
    \label{tab:ohsumed}
    \vspace{-0.3cm}
\end{table}
\begin{figure}[htb]
    \centering
    \setlength{\abovecaptionskip}{0.cm}
    \setlength{\belowcaptionskip}{-0.cm}
    \includegraphics[width=0.46\columnwidth]{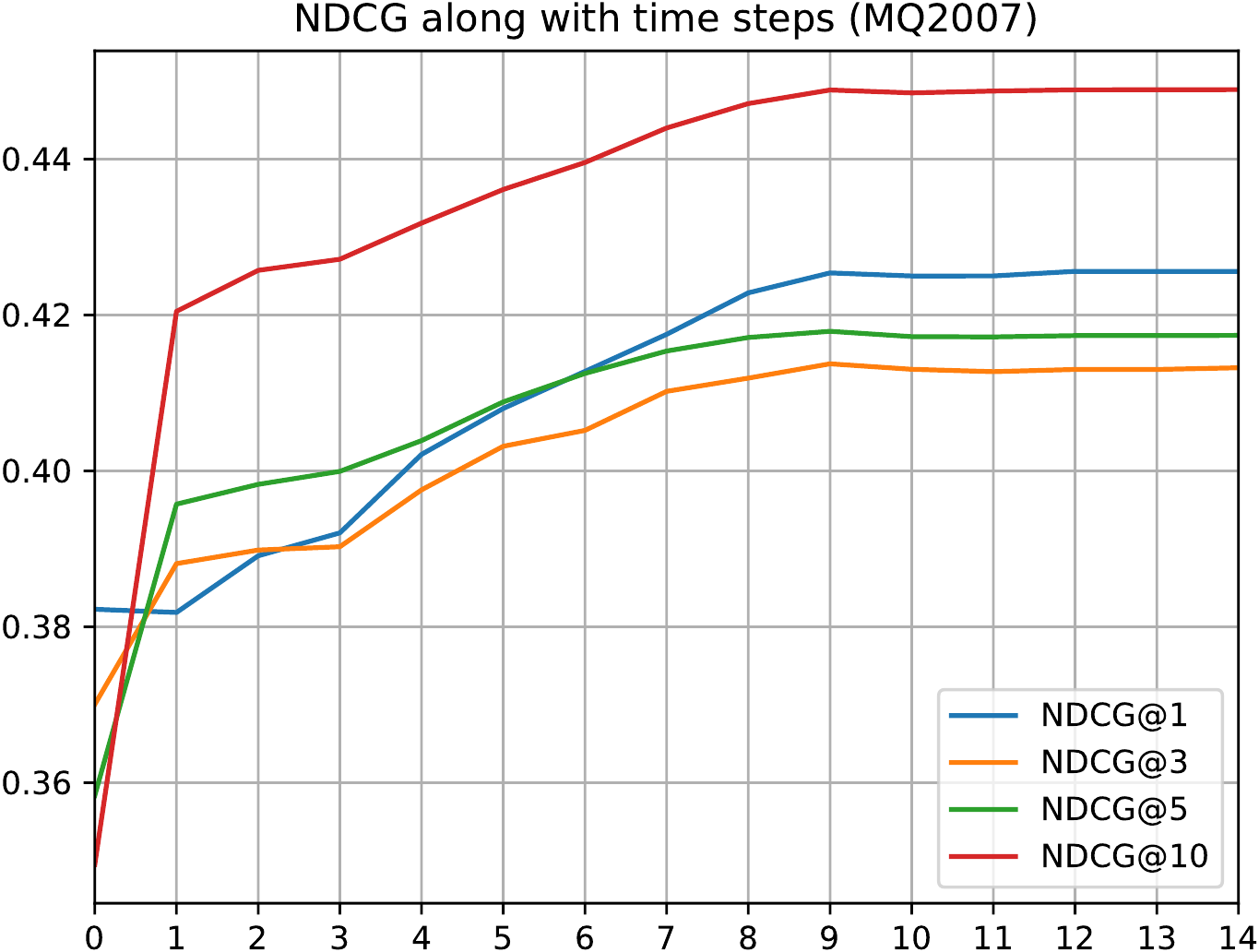}
    \includegraphics[width=0.46\columnwidth]{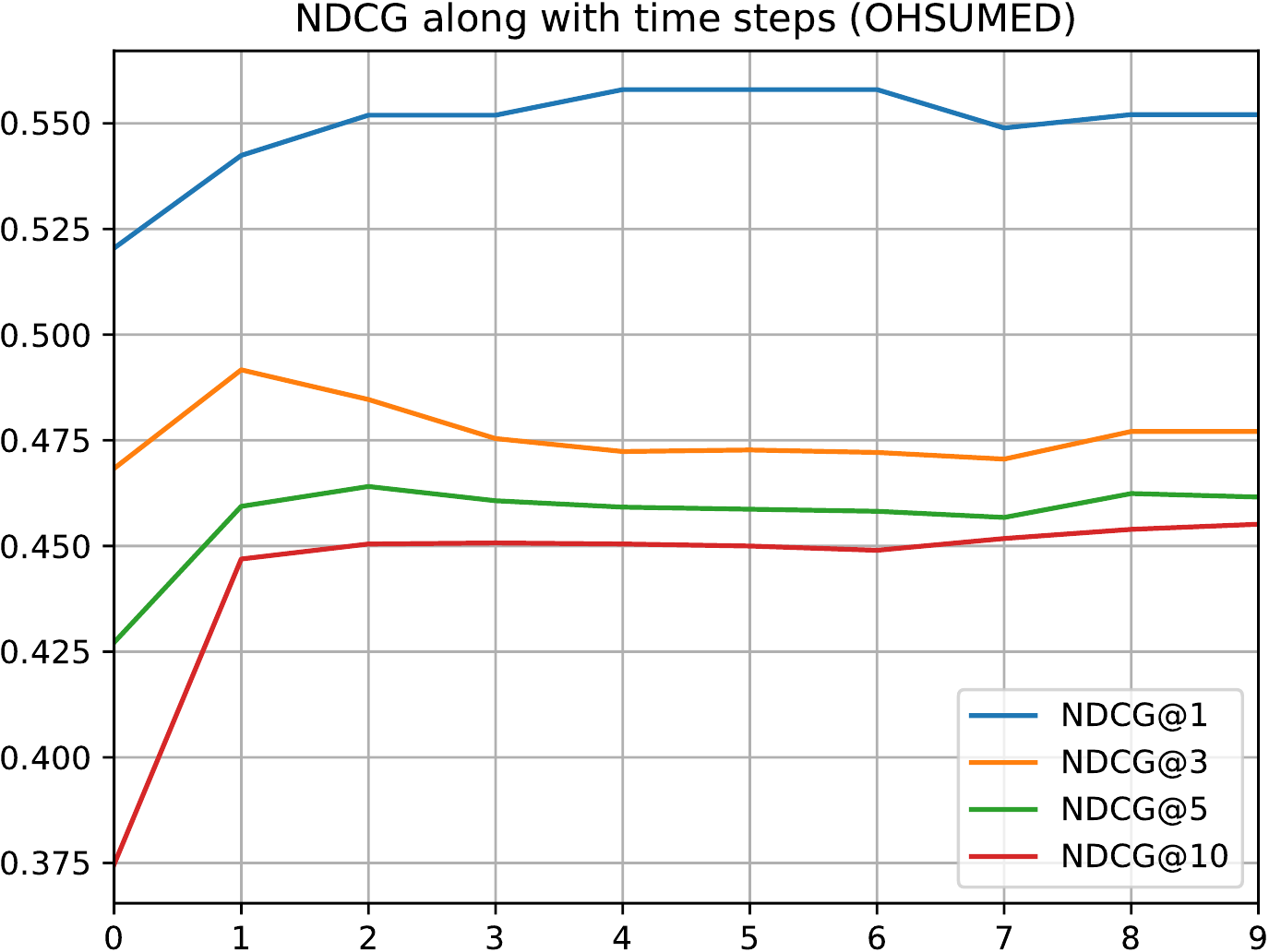}
    \caption{NDCG@1,3,5 and 10 increase along with the time step, demonstrating the interactions (mutual information) among documents help improve the ranking performance.}
    \label{fig:ndcg_along_with_time}
    \vspace{-0.5cm}
\end{figure}

\section{Conclusion}
In this paper, we propose MarlRank, a multi-agent framework for the task of learning to rank. By considering each document as an agent, we effectively integrate the mutual information among documents in generating the final ranking list. The experimental results demonstrate the performance improvement of MarlRank over multiple state-of-the-art baselines and also validate the effectiveness of applying document mutual information in such an interactive way in the ranking task. For the future work, more sophisticated approach should be studied to decide neighbor documents from not only similarity but also diversity, novelty, etc. 

\bibliographystyle{ACM-Reference-Format}
\bibliography{sample-bibliography}

\end{document}